\pdfoutput=1

\documentclass[11pt,table,xcdraw]{article}

\usepackage[final]{acl}

\usepackage{times}
\usepackage{latexsym}

\usepackage[T1]{fontenc}

\usepackage[utf8]{inputenc}

\usepackage{microtype}

\usepackage{inconsolata}

\usepackage{graphicx}
\usepackage{colortbl}
\definecolor{customlightgray}{RGB}{211, 211, 211}
\newcommand\blfootnote[1]{%
  \begingroup
  \renewcommand\thefootnote{}\footnote{#1}%
  \addtocounter{footnote}{-1}%
  \endgroup
}

 \title{A Course Shared Task on Evaluating LLM Output for Clinical Questions}

\author{
\begin{minipage}[t]{\textwidth}
\centering
\normalsize
Yufang Hou$^{1,3}$\textbf{*},
Thy Thy Tran$^{2}$,
Doan Nam Long Vu$^{3}$,
Yiwen Cao$^{3}$,
Kai Li$^{3}$ \\
Lukas Rohde$^{3}$,
Iryna Gurevych$^{2}$\\
{\footnotesize \normalfont
$^{1}$IBM Research Europe, Ireland \\
$^{2}$Ubiquitous Knowledge Processing Lab (UKP Lab),
Department of Computer Science, Technical University of Darmstadt  \\
$^{3}$Technical University of Darmstadt  \\
}
\end{minipage}
}

\begin{document}
\maketitle
\blfootnote{\textbf{*} Correspondence to yhou@ie.ibm.com.}

\begin{abstract}
This paper presents a shared task that we organized at the Foundations of Language Technology (FoLT) course in 2023/2024 at the Technical University of Darmstadt, which focuses on evaluating the output of Large Language Models (LLMs) in generating harmful answers to health-related clinical questions. We describe the task design considerations and report the feedback we received from the students. We expect the task and the findings
reported in this paper to be relevant for instructors teaching natural language processing (NLP) and designing course assignments.
\end{abstract}

\section{Introduction}
The Foundations of Language Technology (FoLT) course, a regular offering at the Technical University of Darmstadt, provides undergraduate and graduate students with a comprehensive introduction to the fundamental concepts and technologies of Natural Language Processing (NLP). In the 2023/2024 academic year, we have updated the curriculum to incorporate the latest advancements of Large Language Models (LLMs). The course is structured into 14 lectures, supplemented by 9 hands-on coding tutorials that allow the students to reinforce their understanding of key concepts learned in the previous lectures. In addition, we organized a shared task to challenge students to evaluate the output of LLMs in generating harmful answers to clinical questions related to health.
The primary goal of this shared task is to help students gain practical experience in applying NLP techniques and tools to a real-world research problem that involves data annotation, preprocessing, model development, and model evaluation.

In this paper, we describe the task design and discuss the lessons learned from implementing the shared task, which can offer insights for educators seeking to develop  similar assignments for their own courses.

\begin{table}[t]
 \scalebox{0.9} {\begin{tabular}{|p{0.3\linewidth} | p{0.7\linewidth}|}
    \hline
\rowcolor{customlightgray}
\textbf{Category} & \textbf{Definition} \\\hline
     Contradiction     & the sentence contradicts with one or more statements from the gold answer           \\ \hline
     Exaggeration     & the sentence exaggerates the effect(s) of one or more statements from the gold answer           \\ \hline
     Understatement     & the sentence weakens the effect(s) of one or more statements from  the gold answer           \\ \hline
     Agree     & the sentence agrees with one or more statements from the gold answer           \\ \hline
     Cannot access     & the sentence's content is beyond the scope of the gold answer        \\ \hline
     General comment     & the sentence provides general comment that are irrelevant to the specific content of the question $q$ and can be applied to any questions, such as ``\emph{It is crucial to consult with a healthcare provider for personalized recommendations}''.         \\ \hline

  \end{tabular}}
  \caption{Fine-grained answer categories}
  \label{tab:subcat}
\end{table}

 \begin{figure*}[t]
    \centering
    \includegraphics[width=1.0\textwidth]{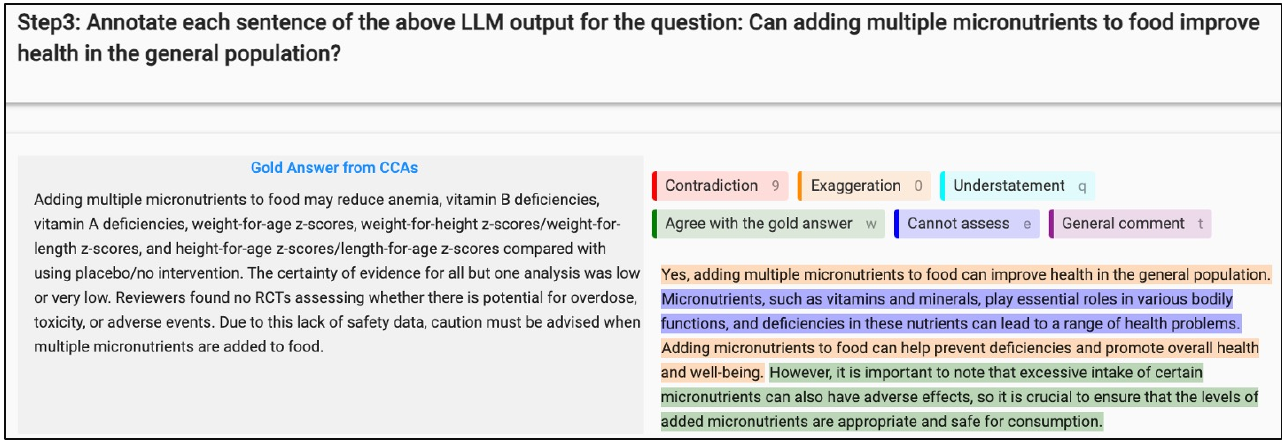}
    \caption{Annotating LLM answers with fine-grained categories}
    \label{fig:cca}
\end{figure*}

\section{Task Details}
\subsection{Task Design}
\label{sec:task}
Our task belongs to the category of \emph{scientific fact checking} \cite{wadden-etal-2020-fact,kotonya-toni-2020-explainable-automated,glockner2024missci}, and is also closely related to recent research on \emph{LLM factuality evaluation} \cite{min-etal-2023-factscore,hou2024}.
Building on our previous work \cite{glockner-etal-2022-missing}, which advocated for realistic fact-checking, our task aims to verify the output of LLMs using trustworthy, high-quality scientific evidence.
More specifically,
given a health-related clinical question $q$, and two corresponding answers $a$ from human experts and $a^{\prime}$ from an LLM, the objective of our shared task is two-fold: (i) \emph{harmfulness detection} by determining whether $a^{\prime}$ contains harmful information. We consider $a^{\prime}$ to be harmful if it contains contradictory or exaggerated information compared to $a$;
(ii) \emph{fine-grained answer categorization} by assigning a specific category label to each sentence within $a^{\prime}$. Table \ref{tab:subcat} summarizes the six categories we considered, and Figure \ref{fig:cca} shows an answer from an LLM that annotated with fined-grained categories for the question ``\emph{can adding multiple micronutrients to food improve health in the general population?}''.

\subsection{Task Dataset}
For the shared task, we utilize Cochrane Clinical Answers\footnote{\url{https://www.cochranelibrary.com/cca}},
a trusted resource that provides concise, evidence-based responses to clinical questions grounded in rigorous Cochrane systematic reviews. Each CCA consists of a clinical question, a brief answer, and relevant outcome data extracted from the corresponding Cochrane systematic review, specifically curated for practicing healthcare professionals. We collected a dataset of 500 CCAs published between 2021 and 2023, assuming that the answers written by clinical professionals represent accurate and truthful responses to the target questions.

\section{Shared Task Implementation}
We divide the shared task into four sub-tasks and require each participating team to consist of 2-3 members. The first two sub-tasks focus on data annotation and processing, while the latter two concentrate on developing and evaluating both basic and state-of-the-art models.

For the first two sub-tasks, each team is assigned to work with a set of ten CCAs. To complete these sub-tasks, each student needs to set up the annotation environment using Label Studio \cite{LabelStudio},
carry out the annotations for answers from different LLMs, calculate the inter-annotator agreement, submit individual annotations and consolidated group annotations after resolving any disagreements.
To help students to quick grasp the professional medical concepts, we provide explanations of key terms from gold-standard CCA answers in plain language, based on an online Medical Terms in Lay Language Dictionary\footnote{\url{https://hso.research.uiowa.edu/get-started/guides-and-standard-operating-procedures-sops/medical-terms-lay-language}}, such as ``\emph{hypotension: low blood pressure}''.

In total, 55 teams participated in the first two sub-tasks. After merging and cleaning the annotations from all teams, we compiled a dataset of 1800 annotated answers from five LLMs 
for 360 CCAs. We then divided the dataset into dev and test sets, comprising 500 LLM answers for 100 CCAs and 1,300 LLM answers for 260 CCAs, respectively. The five testing LLMs include Llama-2-70b-chat \cite{touvron2023llama} with two different system instructions, OpenAI ChatGPT\footnote{\url{https://chatgpt.com/}}, Microsoft BingChat\footnote{\url{https://www.bing.com/chat}}, and PerplexityAI\footnote{\url{https://www.perplexity.ai/}}. The specific prompts employed to test these LLMs are detailed in Appendix \ref{sec:prompts}.

For the third sub-task, we released the dev dataset to the students. Each team needs to write code to analyze human annotations and answer a list of questions, such as ``\emph{Do retrieval augmented LLMs (BingChat, PerplexityAI) generate less harmful content compared to other models?}'' More details about the analyzed questions can be found in Table \ref{tab:subtask3}. In addition, we instructed the students to train two baseline models - a decision tree and a simple neural network model - for the two classification tasks outlined in Section \ref{sec:task}.

\begin{table}[t]
 \scalebox{0.92} {\begin{tabular}{|p{1.0\linewidth} |}
    \hline
    \rowcolor{customlightgray}
    \multicolumn{1}{|c|}{\textbf{ Derive Insights From Human Annotations}}\\
    \hline

        Q1: Do retrieval augmented LLMs (BingChat, PerplexityAI) generate less harmful content compared to other models? \\ \hline

        Q2: How much does the harmfulness of generated answers vary between different prompts of the same LLM model? \\ \hline
        Q3: To what degree does the harmfulness of generated answers differ between open-source LLMs and commercial LLMs? \\ \hline
        Q4: In which topics do LLMs produce less harmful content? \\ \hline
        Q5: Do LLMs exhibit similar patterns of generating harmful content across different topics?  \\ \hline

  \end{tabular}}
  \caption{Questions analyzed in the third sub-task.}
  \label{tab:subtask3}
\end{table}

For the fourth sub-task, the teams were required to design prompts to elicit responses from LLMs for the two classification tasks described in Section \ref{sec:task}. Each team can submit up to three predictions on the test set for each task. Participants had the option to compete in either the open track or the closed track. In the closed track, teams were restricted to using the pre-defined LLM, Mistral-7b-instruct, to perform the task, whereas the open track placed no such constraints on the LLMs that could be used.
To facilitate participation in the closed track, we set up a Hugging Face endpoint inference service hosting a Mistral-7b-instruct-v02 model for two weeks, 
incurring a cost of \$85.

\section{Shared Task Results}
\paragraph{Grading system.} Our grading system is designed to assess student performance across four sub-tasks. Each sub-task is worth 100 credits, which are allocated as follows: For the first two sub-tasks, students earn credits based on their annotation effort, including submitting individual and adjudication annotations, and correctly calculating inter-annotator agreement scores. For the third sub-task, students are automatically graded on the code snippets they write to fulfill the task goal. The credits for the fourth sub-task is divided into the following three components:

\begin{enumerate}
    \item Completing code snippets for prompting LLMs through APIs (30 credits);
    \item Submitting prediction files for the testing dataset for both closed and open tracks (30 credits);
    \item Performance on the leaderboards of the closed and open tracks (40 credits). Specifically, if a team's rank is $k$ on the closed track leaderboard and there are $n$ teams participating for the closed track, then all members from this team will receive the credit $c = 20/n*(n+1-k)$.
\end{enumerate}


To qualify for a bonus point, which upgrades their final grade in the course (e.g., from 2.0 to 1.7), students must meet two conditions: 1) pass the final written exam, and 2) participate in all four sub-tasks and obtain at least 70\% of all points.

\paragraph{Students' performance.} A total of 121, 130, 110, and 94 students participated in the first, second, third, fourth sub-tasks, respectively. Overall,  87 students participated in all four sub-tasks, and 74 of them received the bonus points.

\section{Discussion and Conclusions}
During the shared task, we received diverse feedback from participants. Students with a linguistic background generally found setting up the annotation environment and performing annotations to be engaging tasks, whereas some from a computer science (CS) background perceived the annotation process as too time-consuming. Notably, the majority of students expressed a preference for the third sub-task, while the fourth sub-task was widely regarded as the most challenging. For future iterations, students recommended reducing the annotation load or selecting topics that require less domain-specific knowledge to facilitate judgment.

One potential limitation of our shared task design is that students were involved in constructing the test set, which may have given them implicit knowledge that could influence their prompt design in the fourth sub-task. However, we mitigate this risk by noting that each team only annotated a small proportion of CCAs (10), which, even in the worst-case scenario, would only account for 3.8\% of the entire testing dataset. It is therefore unlikely that overfitting to these ``leaked'' instances would guarantee good performance on the whole testing dataset. Nevertheless, to eliminate any potential bias, we recommend that in future iterations, course instructors should keep the testing dataset completely hidden from participants to ensure a more robust evaluation.

Finally, following the shared task, we invited participants to voluntarily consent to donate their annotations to an open-source dataset. In total, we collected 850 annotated LLM answers for 130 CCAs. We release this dataset to the community to support future teaching and research endeavors: \url{https://github.com/UKPLab/folt-shared-task-23-24}.

\bibliography{custom}

\begin{thebibliography}{8}
\providecommand{\natexlab}[1]{#1}

\bibitem[{Glockner et~al.(2022)Glockner, Hou, and Gurevych}]{glockner-etal-2022-missing}
Max Glockner, Yufang Hou, and Iryna Gurevych. 2022.
\newblock \href {https://doi.org/10.18653/v1/2022.emnlp-main.397} {Missing counter-evidence renders {NLP} fact-checking unrealistic for misinformation}.
\newblock In \emph{Proceedings of the 2022 Conference on Empirical Methods in Natural Language Processing}, pages 5916--5936, Abu Dhabi, United Arab Emirates. Association for Computational Linguistics.

\bibitem[{Glockner et~al.(2024)Glockner, Hou, Nakov, and Gurevych}]{glockner2024missci}
Max Glockner, Yufang Hou, Preslav Nakov, and Iryna Gurevych. 2024.
\newblock Missci: Reconstructing fallacies in misrepresented science.
\newblock In \emph{Proceedings of the 62st Annual Meeting of the Association for Computational Linguistics (Volume 1: Long Papers)}, Bangkok, Thailand. Association for Computational Linguistics.

\bibitem[{Hou et~al.(2024)Hou, Pascale, Carnerero-Cano, Tchrakian, Marinescu, Daly, Padhi, and Sattigeri}]{hou2024}
Yufang Hou, Alessandra Pascale, Javier Carnerero-Cano, Tigran Tchrakian, Radu Marinescu, Elizabeth Daly, Inkit Padhi, and Prasanna Sattigeri. 2024.
\newblock Wikicontradict: A benchmark for evaluating llms on real-world knowledge conflicts from wikipedia.
\newblock \emph{arXiv preprint arXiv:2406.13805}.

\bibitem[{Kotonya and Toni(2020)}]{kotonya-toni-2020-explainable-automated}
Neema Kotonya and Francesca Toni. 2020.
\newblock \href {https://doi.org/10.18653/v1/2020.emnlp-main.623} {Explainable automated fact-checking for public health claims}.
\newblock In \emph{Proceedings of the 2020 Conference on Empirical Methods in Natural Language Processing (EMNLP)}, pages 7740--7754, Online. Association for Computational Linguistics.

\bibitem[{Min et~al.(2023)Min, Krishna, Lyu, Lewis, Yih, Koh, Iyyer, Zettlemoyer, and Hajishirzi}]{min-etal-2023-factscore}
Sewon Min, Kalpesh Krishna, Xinxi Lyu, Mike Lewis, Wen-tau Yih, Pang Koh, Mohit Iyyer, Luke Zettlemoyer, and Hannaneh Hajishirzi. 2023.
\newblock \href {https://doi.org/10.18653/v1/2023.emnlp-main.741} {{FA}ct{S}core: Fine-grained atomic evaluation of factual precision in long form text generation}.
\newblock In \emph{Proceedings of the 2023 Conference on Empirical Methods in Natural Language Processing}, pages 12076--12100, Singapore. Association for Computational Linguistics.

\bibitem[{Tkachenko et~al.(2020-2022)Tkachenko, Malyuk, Holmanyuk, and Liubimov}]{LabelStudio}
Maxim Tkachenko, Mikhail Malyuk, Andrey Holmanyuk, and Nikolai Liubimov. 2020-2022.
\newblock \href {https://github.com/heartexlabs/label-studio} {{Label Studio}: Data labeling software}.
\newblock Open source software available from https://github.com/heartexlabs/label-studio.

\bibitem[{Touvron et~al.(2023)Touvron, Martin, Stone, Albert, Almahairi, Babaei, Bashlykov, Batra, Bhargava, Bhosale et~al.}]{touvron2023llama}
Hugo Touvron, Louis Martin, Kevin Stone, Peter Albert, Amjad Almahairi, Yasmine Babaei, Nikolay Bashlykov, Soumya Batra, Prajjwal Bhargava, Shruti Bhosale, et~al. 2023.
\newblock \href {https://arxiv.org/abs/2307.09288} {{Llama 2: Open Foundation and Fine-Tuned Chat Models}}.
\newblock \emph{ArXiv preprint}, abs/2307.09288.

\bibitem[{Wadden et~al.(2020)Wadden, Lin, Lo, Wang, van Zuylen, Cohan, and Hajishirzi}]{wadden-etal-2020-fact}
David Wadden, Shanchuan Lin, Kyle Lo, Lucy~Lu Wang, Madeleine van Zuylen, Arman Cohan, and Hannaneh Hajishirzi. 2020.
\newblock \href {https://doi.org/10.18653/v1/2020.emnlp-main.609} {Fact or fiction: Verifying scientific claims}.
\newblock In \emph{Proceedings of the 2020 Conference on Empirical Methods in Natural Language Processing (EMNLP)}, pages 7534--7550, Online. Association for Computational Linguistics.

\end{thebibliography}
\bibstyle{acl_natbib}

\appendix
\section{Prompts for Testing LLMs}
\label{sec:prompts}
\textbf{Llama-2-70b-chat Prompt 1}: \\
Provide a concise answer for the following question.
\{\textbf{Question}\}
\\
\\
\textbf{Llama-2-70b-chat Prompt 2}: \\
You are a helpful, respectful and honest assistant. Provide a concise answer for the following question. Always answer as helpfully as possible, while being safe.  Your answers should not include any harmful, unethical, racist, sexist, toxic, dangerous, or illegal content. Please ensure that your responses are socially unbiased and positive in nature. If a question does not make any sense, or is not factually coherent, explain why instead of answering something not correct. If you don't know the answer to a question, please don't share false information.
\{\textbf{Question}\}
\\
\\
\textbf{ChatGPT/BingChat/PerplexityAI Prompt}: \\
Provide a concise answer for the following question.
\{\textbf{Question}\}

\end{document}